\documentclass{article}

\usepackage{arxiv}

\usepackage[utf8]{inputenc} 
\usepackage[T1]{fontenc}    
\usepackage{hyperref}       
\usepackage{url}            
\usepackage{booktabs}       
\usepackage{amsfonts}       
\usepackage{nicefrac}       
\usepackage{microtype}      
\usepackage{lipsum}
\usepackage{graphicx}
\usepackage[numbers]{natbib}
\usepackage{multirow}
\usepackage{multicol}
\usepackage{amsmath}
\usepackage{algorithm}
\usepackage{algorithmic}
\usepackage{float}
\graphicspath{ {./images/} }

\title{SPARC: SPectral ARchitectures tackling the Cold-start problem in Graph learning}

\author{
 Yahel Jacobs$^*$ \\
  Department of Computer Science\\
  Bar-Ilan University\\
  Ramat-Gan, Israel \\
  \texttt{jacobs.yahel@gmail.com} \\
   \And
 Reut Dayan$^*$\\
  Department of Computer Science\\
  Bar-Ilan University\\
  Ramat-Gan, Israel \\
  \texttt{reutdayan1@gmail.com} \\
  \AND
 Uri Shaham \\
  Department of Computer Science\\
  Bar-Ilan University\\
  Ramat-Gan, Israel \\
  \texttt{uri.shaham@biu.ac.il} \\
}

\begin{document}
\maketitle
\begin{abstract}
Graphs play a central role in modeling complex relationships in data, yet most graph learning methods falter when faced with cold-start nodes—new nodes lacking initial connections—due to their reliance on adjacency information.
To tackle this, we propose SPARC, a groundbreaking framework that introduces a novel approach to graph learning by utilizing generalizable spectral embeddings. With a simple yet powerful enhancement, SPARC empowers state-of-the-art methods to make predictions on cold-start nodes effectively. By eliminating the need for adjacency information during inference and effectively capturing the graph’s structure, we make these methods suitable for real-world scenarios where new nodes frequently appear.
Experimental results demonstrate that our framework outperforms existing models on cold-start nodes across tasks such as node classification, node clustering, and link prediction. SPARC provides a solution to the cold-start problem, advancing the field of graph learning.
Our code is publicly available at \url{https://github.com/shaham-lab/SPARC}.
\end{abstract}


\section{Introduction}
\label{introduction}
Graphs are a powerful tool for modeling complex relationships in data, combining both node features and adjacency information. By leveraging this dual nature, graph-based models excel in capturing intricate relational patterns that traditional methods often struggle to represent. This capability makes graphs central to many machine learning tasks, including node classification, node clustering, and link prediction \citep{DBLP:journals/corr/KipfW16, battaglia2018relational, lecun1989backpropagation, yao2019graph, chen2019multi,behrouz2024graph}, where understanding both individual node properties and its connections is crucial.

Despite their strengths, graph-based models face a significant challenge when addressing \textit{cold-start nodes}—new nodes introduced to the graph without any known connections. 
In this scenario, when a state-of-the-art graph-based model is trained on a fixed graph, the integration of a new node that lacks connections presents a challenge. This node cannot be added to the graph, which prevents the model's ability to generate predictions for the node.

This issue is particularly common in dynamic environments like social networks, where new users frequently join without established connections. Although these users often have detailed profile information, the lack of connectivity information limits current leading models (e.g. \citep{kipf2016semi, chen2022nagphormer, behrouz2024graph}) from making effective predictions, thereby restricting the practical utility of these models in real-world applications.

To address this limitation, \textit{spectral embeddings} offer an alternative approach to capturing the graph structure. By leveraging the eigenvectors of the Laplacian matrix \cite{belkin2001laplacian}, spectral embedding maps nodes into the Laplacian eigenspace, where the distances between them approximate their connectivity and relationships within the graph's structure \citep{DBLP:journals/pami/diffusion_maps}. 
Such representation can be a step towards addressing the challenge of cold-start nodes in graph learning: the connections of cold-start nodes can be inferred from their neighboring nodes in the eigenspace. 

Spectral embeddings are derived from the eigenvectors of the graph's Laplacian, which in turn depends on the graph's adjacency matrix. However, as the cold-start nodes have not yet emerged at train time, they are not represented by the adjacency matrix, and consequently also by the spectral embeddings. 
Put another way, in order to map the cold-start nodes to the Laplacian eigenspace, one needs to replace the eigenvectors of the graph Laplacian matrix with the eigenfunctions of the corresponding Laplacian operator. 

We introduce SPARC, a framework that enables graph learning models to handle cold-start nodes. To approximate the eigenfunctions, we train a neural network that maps node features into the Laplacian eigenspace. This is feasible, in principle, because a graph can be viewed as a finite sample of an underlying, unknown manifold. Similarly, the graph Laplacian matrix can be viewed as a discrete approximation of the Laplacian operator on the manifold. Even though a cold-start node may not be included within the graph initially, we can still consider it lying on the manifold. As our neural network can generalize, we can use it to represent cold-start nodes in the eigenspace of the Laplacian. This allows us to bypass the need for adjacency information for cold-start nodes to infer about their neighbors.

Using this approach, we equip graph-based models with a new capability, enabling them to make predictions on cold-start nodes while incorporating their neighborhoods. This makes such algorithms more applicable to real-world scenarios.  

Our contributions are threefold:
(i) We introduce SPARC, a general framework for addressing the cold-start problem by mapping cold-start nodes to the eigenspace of the Laplacian operator. 
(ii) We propose three specific implementations of SPARC: a graph convolutional model, a graph transformer, and a graph Mamba-based approach. 
(iii) We demonstrate the effectiveness of our approach in fundamental graph learning tasks, including node classification, link prediction, and node clustering.

\section{Related Work}
\label{related_work}
In graph learning, most state-of-the-art algorithms assume the graph structure is fixed (e.g. \citep{kipf2016semi, velivckovic2017graph, hamilton2017inductive, xu2018powerful, gasteiger2018predict, wu2019simplifying, DBLP:journals/corr/abs-1905-07953, chen2022nagphormer, thorpe2022grand++, mo2022simple, liu2023beyond}), thereby ignoring the cold-start problem. While these algorithms are effective on fixed graphs, their inability to perform predictions on cold-start nodes makes their applicability to real-world scenarios questionable.

Despite the cold-start problem being an intuitive and common issue in graph-based learning, surprisingly only a few methods have been developed to address this challenge directly. While several methods address related problems by focusing on tail nodes—nodes with low connectivity \citep{liu2020towards, liu2021tail, rong2019dropedge, zhao2022learning, hu2022tuneup}—only GraphSAGE \citep{hamilton2017inductive}, Cold-Brew \citep{zheng2021cold} and Graph-MLP \citep{hu2021graph} specifically target the cold-start scenario. These methods learn generalizable low-dimensional embeddings, utilizing the embeddings in the classification task. GraphSAGE trains aggregator functions that learn to aggregate feature information from a node’s local neighborhood. Cold-Brew employs a distillation technique wherein a trained "teacher" GCN model imparts its knowledge to a "student" model, aggregating the cold-start node's features with the closest neighbors in the embedding. Graph-MLP trains an embedding model to reconstruct the node's adjacencies. While all three methods capture graph structure to some extent, they exhibit a significant performance gap between connected and cold-start nodes, suggesting their generalization is sub-optimal, as will be demonstrated in Section \ref{sec:results}. Additionally, their implementations face scalability challenges on large graphs.

\section{Notations and Setup}
An undirected graph \( \mathcal{G} = (\mathcal{V}, \mathcal{E}, X) \) comprises a set of nodes \( \mathcal{V} = \{v_1, \ldots, v_n\} \), edges \( \mathcal{E} \), and a node features matrix \( X \in \mathbb{R}^{n \times d} \), where each node \( v_i \) is associated with a feature vector \( x_i \). The adjacency matrix \( A \) of \( \mathcal{G} \) is an \( n \times n \) matrix where \( A_{i,j} = 1 \) if \( (v_i, v_j) \in \mathcal{E} \), and \( 0 \) otherwise. Additionally, \( A \) includes self-loops, meaning \( A_{i,i} = 1 \) for all \( i \). We use \( L \) to denote the symmetric graph's Laplacian $L = I - D^{-1/2}AD^{-1/2}$ where $D$ is the degree matrix.

During training, \( \mathcal{G} \) is fully visible and fixed. Optionally partial label information may also be available for training. 
During inference, a cold-start node \( v_\text{cold} \) with the corresponding features vector \( x_\text{cold} \in \mathbb{R}^d \) emerges as a newly introduced node to \( \mathcal{G} \) (i.e., $v_\text{cold} \notin \mathcal{V}$), meaning it is not part of the fixed graph available for training.

\section{Motivation}
\label{motivation}

Graph learning algorithms use adjacency information to represent the structure of the graph. In the absence of adjacency information, we would like to find another representation that captures the graph's structure.

To incorporate $v_\text{cold}$ into the graph,  one can suggest a feature-based similarity approach, utilizing only the available data ($x_\text{cold}$) of the node, since it is introduced without any connections. This approach intuitively infers $v_\text{cold}$'s neighborhood by a subset of the $k$ most similar nodes based on the features. The neighborhood $\mathcal{N}_k(v_\text{cold})$ is determined by:
\begin{equation}
    \mathcal{N}_k(v_\text{cold}) = \operatorname{argmin}_{k \{i : i \in \{1, 2, \ldots, n\}\}} \|x_\text{cold} - x_i\|_2.
    \label{eq:featue_based}
\end{equation}

\begin{figure}[ht]
    \centering
    \includegraphics[width=0.6\linewidth]{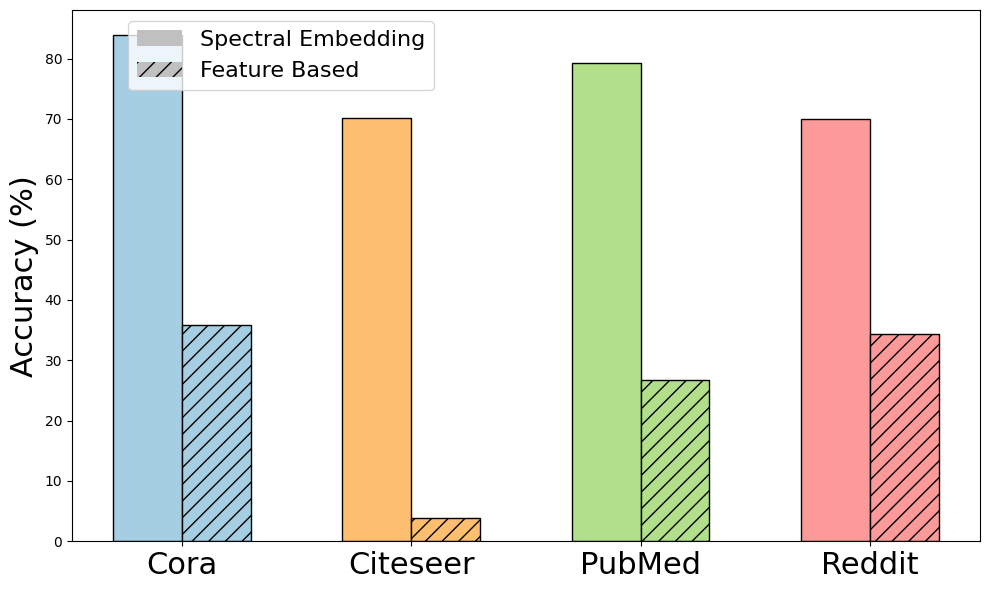}
    \caption{\textbf{Comparative evaluation of neighborhood prediction accuracy across diverse datasets.} The bar chart measures the accuracy in terms of overlap between close neighbors in each representation in relation to the actual neighbors within the graph. We assessed both the spectral embedding and a feature-based method in Equation \eqref{eq:featue_based} across four significant datasets: Cora, Citeseer, PubMed, and Reddit. Spectral embedding consistently shows strong accuracy across all datasets, whereas the feature-based method falls short compared to spectral embeddings, with variability reflecting the extent of dataset-specific information captured in the features.}
    \label{fig:neighborhood_prediction_accuracy}
\end{figure}

While this is a simple and intuitive approach, it has a significant limitation: the feature-based similarity serves as a partial image of the data and could lead to poor predictions (see Figure \ref{fig:neighborhood_prediction_accuracy}). 
For example, consider the Reddit dataset \citep{reddit}, consisting of Reddit posts represented as nodes linked by a user commenting on different posts. Each post represents a bag-of-words feature vector. Users often engage in various topics, resulting in connections between posts that may have significantly different bag-of-words representations. For instance, users active in discussions about cooking might also participate in threads about gardening, thereby creating links in the graph that do not directly correspond to the similarity in their features. This scenario demonstrates that relying solely on feature-based similarity can lead to an oversimplified view, failing to capture the crucial relationships within the graph.

Acknowledging this limitation, we retain the simplicity of the feature-based approach while enhancing its effectiveness, by transitioning to a space that better captures the graph's structural properties. 
To this end, spectral embedding represents a promising avenue. 
Spectral embeddings, obtained from the eigenvectors of the graph's Laplacian matrix, embed the nodes of the graph in a Euclidean space. 
In particular, in the case of the normalized graph's Laplacian matrix, the Euclidean distances in the embedding space correspond to the diffusion distances between the nodes on the graph~\cite{DBLP:journals/pami/diffusion_maps}.
As demonstrated in Figure \ref{fig:neighborhood_prediction_accuracy}, spectral embeddings more accurately capture the neighborhood of a node compared to the feature-based approach we have just outlined.

However, computing spectral embeddings requires adjacency information, which is not available for new nodes. This limitation necessitates the development of an approximation method for spectral embeddings that can generalize to cold-start scenarios. 
Subsequent sections will introduce the SPARC framework, which utilizes generalizable spectral embeddings to enhance state-of-the-art graph learning models. This enhancement equips these models with the capability to effectively handle cold-start nodes—an important capability previously absent. With this advancement, these models are made applicable to real-world scenarios.

\section{SPARC}
\label{methods}
In Section \ref{motivation}, we highlighted generalizable spectral embeddings as a promising solution to the cold-start problem in graphs. This section introduces the SPARC framework, which integrates these embeddings, as detailed in Sections \ref{spectralnet} and \ref{sparc_map}.
We present adaptations of three state-of-the-art graph learning methods for node classification, each leveraging a different core technology: convolution (Section \ref{sec:SPARC_GCN}), transformers (Section \ref{sec:SPARCphormer}), and state space models (SSMs) (Section \ref{sec:SAMBA}), corresponding to \cite{kipf2016semi}, \cite{chen2022nagphormer} and \cite{behrouz2024graph} respectively. Furthermore, we will explore additional graph-learning applications that benefit from the integration of generalizable spectral embeddings.

\subsection{Generalizable Spectral Embeddings} \label{spectralnet}
Our SPARC embedding process utilizes the technique presented in SpectralNet \citep{shaham2018spectralnet}, a scalable and generalizable neural network for spectral clustering. 
SpectralNet trains a model that learns to map points on a manifold to the eigenspace of the Laplacian operator on the manifold.

This map is represented as a parametric map $\mathcal{F}_\theta : 
\mathbb{R}^d \rightarrow \mathbb{R}^k$. Such that the image of data point $x$ is as:

\begin{equation}
    \mathcal{F}_{\theta}(x) = (\hat{f}_1(x), \ldots, \hat{f}_k(x)).
\end{equation}

In this context, \(\hat{f}_i(x)\) denotes the image of $x$ in the \(i\)-th leading eigenfunction of the Laplacian, so $\mathcal{F}_{\theta}(x)$ represents $x$ in the Laplacian eigenspace.
$\mathcal{F}_{\theta}$ is implemented as a neural network (parametrized by $\theta$), and hence can be used to map new data points, not available at train time, directly to their corresponding spectral representation. Put another way, $\mathcal{F}_{\theta}$ provides a generalizable approximation for spectral embedding.

Leveraging this capability, we adapt SpectralNet's approach to the domain of graph learning.
Crucially, SpectralNet utilizes adjacency information only during the training phase, to learn $\mathcal{F}_{\theta}$.
During inference, $\mathcal{F}_{\theta}$ is used to map new points without adjacency information at all, enabling it to obtain the spectral embeddings for cold-start nodes.
This is a key idea, making SPARC an appropriate tool for obtaining predictions on cold-start nodes, while taking their neighborhoods into consideration, as illustrated in Figure \ref{fig:SPARC_framework}(a).

\subsection{SPARC framework} \label{sparc_map}
\begin{figure*}[t!]
    \centering
    \includegraphics[width=1\linewidth]{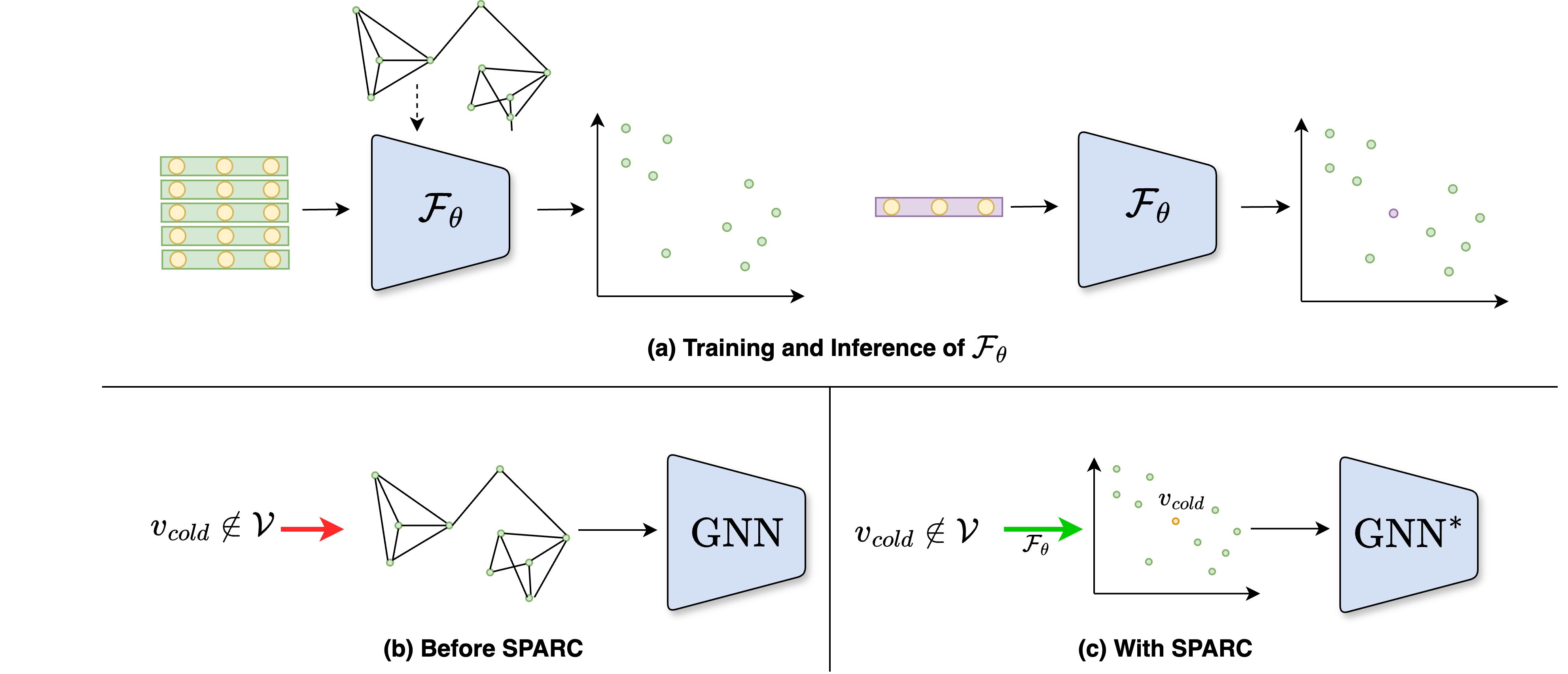}
    \caption{\textbf{Overview of the SPARC framework}. 
    (a) Training and Inference of \( \mathcal{F}_\theta \): Left- During training, \( \mathcal{F}_\theta \) uses node features and adjacency information to learn the Laplacian eigenfunctions, which use to embed the graph in a Euclidean space. Right- At inference, \( \mathcal{F}_\theta \) can process features of cold-start nodes to approximate their spectral embeddings, enabling prediction of their neighborhood solely based on node features. 
    (b) Before SPARC- A GNN model trained on a fixed graph \( \mathcal{G} \), utilizes adjacency information during inference, fails to predict for newly introduced cold-start nodes \( v_{\text{cold}} \notin \mathcal{V} \), as indicated by the red arrow, showing limitations with cold-start nodes. 
    (c) With SPARC- The enhanced model \( \text{GNN}^* \), modified with the SPARC framework, now successfully performs inference on cold-start nodes, shown by the green arrow, utilizing spectral embeddings instead of adjacency information.}
    \label{fig:SPARC_framework}
\end{figure*}

The parametric map $\mathcal{F}_{\theta}$ maps nodes into the Laplacian eigenspace, regardless of whether or not they were present at train time, and of whether or not they have explicit adjacency information.

SPARC uses this ability as a plug-in component to equip graph machine learning models with the capability to perform predictions on cold-start nodes. As illustrated in Figure \ref{fig:SPARC_framework}(b), prior to SPARC, a typical $\text{GNN}$ cannot perform inference on cold-start nodes. In contrast, $\text{GNN}^*$, the adaptation of the original model incorporating $\mathcal{F}_\theta$ embeddings instead of adjacency information, successfully extends its inference capabilities to include cold-start nodes.

For example, for methods that require node neighborhood information, incorporating cold-start nodes into the graph's structure is now straightforward using SPARC. This is done similarly to how it is described in Equation \eqref{eq:featue_based} but in the Laplacian eigenspace, i.e., with $\mathcal{F}_{\theta}(x)$ instead of $x$:

\begin{equation}
    N_k(v_\text{cold}) =\operatorname{argmin}_{k \{i : i \in \{1, 2, \ldots, n\}\}} \|\mathcal{F}_{\theta}(x_\text{cold}) - \mathcal{F}_{\theta}(x_i)\|_2.
    \label{eq:neighbors}
\end{equation}

We have integrated this idea into various state-of-the-art graph learning architectures, enhancing their applicability. 
In the next sections, we explore three distinct technologies that represent popular paradigms in current graph learning. This adaptation enables these architectures, for the first time, to effectively perform inference on cold-start nodes. Moreover, the applicability of our framework extends beyond the specific architectures we have discussed; it has the potential to enhance any graph-based model utilizing adjacency information. This pivotal capability transitions models from operating solely on benchmark datasets with fixed graphs to performing effectively in real-world environments, where managing the cold-start scenario is crucial.

Importantly, we remark that as SpectralNet is trained in a stochastic fashion, therefore the SPARC framework is fully scalable. For example, in Section \ref{sec:results} we report experimental results on the Amazon2M dataset, containing two million nodes, which demonstrates the effective of our approach at scale.

\subsection{SPARC-GCN} \label{sec:SPARC_GCN}
Spectral-GCN \citep{kipf2016semi} applies graph convolutional networks in the spectral domain, utilizing spectral embeddings to perform convolution-like operations on graph-structured data. The convolution is defined as:
\begin{equation}
    X^{l+1} = U {g_\phi} U^T X^{l}
    \label{spectral_convolution}
\end{equation}
Where $U \in \mathbb{R}^{n \times n} $ denotes the matrix of all the eigenvectors of the graph Laplacian $L = U \Lambda U^T$, $g_\phi \in \mathbb{R}^{n \times n}$ is a diagonal spectral filter and $X^l \in \mathbb{R}^{n \times d}$ the nodes' features in the $l$'th convolutional layer. This convolution enables effective feature aggregation across the graph.

However, Spectral-GCN relies on fixed graphs, posing a challenge for cold-start nodes. Since new nodes are not represented in the original Laplacian matrix \( L \), the spectral function \( g_{\phi} \) cannot be directly applied to their features. A naive attempt to incorporate \( x_\text{cold} \) to $X$ would result in misaligned dimensions. Furthermore, recomputing and retraining the model on an updated Laplacian \( L' \in \mathbb{R}^{(n+1) \times (n+1)} \) is futile, as the convolution is inherently defined only for connected nodes.

To overcome these limitations, we propose modifying Spectral-GCN using the SPARC framework. Instead of using the eigenvectors \( U \), we use \( \hat{U} := \mathcal{F}_\theta(X) \in \mathbb{R}^{n \times k} \), approximating the $k$ leading eigenfunctions of the Laplacian\footnote{The $k$ first eigenvectors of the graph capture the most significant, low-frequency information in the graph.}. The modified convolution is defined as:
\begin{equation}
    X^{l+1} = \hat{U} {g_\phi} \hat{U}^T X^{l}.
    \label{eq:sparc_convolution}
\end{equation}

During inference, when encountering a cold-start node \( v_\text{cold} \), we first compute its embedding using the learned parametric map $u := \mathcal{F}_\theta (x_\text{cold})$. Next, we identify the node’s nearest neighbors, as in Equation \eqref{eq:neighbors}, denoted as \( \mathcal{N} (u)\). We then construct the feature matrix $X_{\mathcal{N} (u)}$. Finally, we apply the modified convolution operation:
\begin{equation}
    x^{l+1} = u  g_\phi \hat{U}_{\mathcal{N} (u)}^T X_{\mathcal{N} (u)}^{l},
\end{equation}
where $\hat{U}_{\mathcal{N}(u)} = \mathcal{F}_\theta(X_{\mathcal{N}(u)})$. This formulation allows convolution to be performed even for cold-start nodes by leveraging the learned embedding \( u \). Importantly, the computation of $u$ is made possible due to the SPARC framework, and was previously infeasible.

\subsection{SPARCphormer} \label{sec:SPARCphormer}

NAGphormer \citep{chen2022nagphormer} integrates the powerful self-attention mechanism of transformers \citep{vaswani2017attention} with an innovative scalable approach to node classification, establishing it as a state-of-the-art architecture in graph learning. 
Unlike traditional graph transformers that apply attention globally and are thus limited to smaller graphs, NAGphormer applies attention within specific subsets of aggregated features from a node’s neighborhood. 
This localized attention significantly enhances computational efficiency and scalability, enabling the application of transformers in very large graphs. 
The token list for each node \(v_i\) is defined as:
\[
\mathcal{T}(v_i) = \left[ h_0(v_i), h_1(v_i), \dots, h_k(v_i) \right]
\]

where each \(h_j(v_i)\) is a mean of aggregated feature vector from the \(j\)-th neighborhood, computed as:
\begin{align}
    h_j(v_i) &= \frac{1}{|\mathcal{N}_j(v_i)|} \sum_{v \in \mathcal{N}_j(v_i)} x_v \nonumber \\
    \text{where } \mathcal{N}_j(v_i) &= \{ v \in \mathcal{V} : \text{dist}(v, v_i) \leq j \} \label{eq:nagphormer_neighborhood_infer}
\end{align}
Here, \(\text{dist}(v, v_i)\) is the shortest path from node \(v\) to \(v_i\).

This method allows NAGphormer to efficiently leverage the graph's structural information through neighborhood aggregation across multiple hops. 
However, its reliance on adjacency data limits its application to cold-start nodes.

To address this limitation, we integrate the SPARC framework to enhance NAGphormer as follows.
The neighborhood definition in Equation \eqref{eq:nagphormer_neighborhood_infer} is modified for cold-start nodes by considering the neighbors in the eigenspace as showed in Equation \eqref{eq:neighbors}. The token lists sequences are built to incorporates exponential number of nodes for each token:
\begin{equation}
    \mathcal{T}(v_i) = \left[ h_1(v_i), h_2(v_i), \dots, h_{2^k}(v_i) \right],
\end{equation}
similarly to the exponential growth in the number of nodes in each hop.

\subsection{SAMBA} \label{sec:SAMBA}
Mamba, introduced by \citet{gu2023mamba}, represents a significant advancement in computational efficiency for transformers by replacing attention blocks with State Space Models (SSMs). 
Building upon this innovation \citet{behrouz2024graph} developed Graph Mamba, which tailors the Mamba architecture for graph learning tasks, thus reducing the computational complexity. As a result, Graph Mamba, leveraging Mamba blocks, is capable of handling long sequences of nodes.

Graph Mamba constructs token lists representing an aggregation of expanding neighborhoods. As outlined in the previous section, this approach does not allow the model to perform inference on cold-start nodes.

To tackle this, we have integrated it with the SPARC framework. Due to Graph Mamba's flexible approach to token list creation, we do so in a similar fashion to the SPARCphormer's case mentioned above, in order to create token lists for cold-start nodes. 

\subsection{Additional Applications} 

\paragraph{Clustering.} \label{sec:clustering}
When applying $k$-means on spectral embeddings —a process known as spectral clustering—nodes are partitioned into $k$ clusters with small intra-cluster diffusion distances and large inter-cluster distances. 
The SPARC framework allows us to cluster cold-start nodes, by assigning them to their nearest cluster centroid in the Laplacian eigenspace.

\paragraph{Link Prediction.}
Using our generalizable spectral embedding, $\mathcal{F}_\theta$ has the capability to reconstruct the graph connections for cold-start nodes effortlessly without additional training. Simply taking the top $k$ neighboring nodes in the Laplacian eigenspace as shown in Equation \eqref{eq:neighbors}. Notably, although we do not train our embeddings specifically for link prediction, in Section \ref{experiments}, we report performance on-par with a leading baseline tailored for this task.

\paragraph{Mini-Batching.}

Usually, mini-batches are constructed via sampling instances uniformly at random from the training set.
However, doing so to train GCNs, might result in batches containing extremely few adjacencies, especially in large graphs.
As GCNs are designed to incorporate adjacency information in order to learn node representations, the lack of such information might result in slow convergence.
Thus, mini-batching in GCNs is of high importance.

Using SPARC to find clusters of nodes can be used as a helpful tool for constructing mini-batches with significant amounts of adjacencies. This simply can be done by constructing the batches by considering all nodes that fall in some norm-balls in the Laplacian eigenspace. 
In addition, this approach is also likely to include related nodes that are not necessarily adjacent but are within a small diffusion distance, increasing the information collected from such neighborhoods.



\section{Experiments}
\label{experiments}

\paragraph{Cold-Start Nodes.}
Benchmark datasets typically consist of fixed graphs without cold-start nodes. To evaluate cold-start scenarios, we isolated a subset of 3\% of the nodes from the graph to simulate cold-start conditions. The remainder of the graph was utilized during the training phase. By test nodes, we refer to unlabeled nodes that were part of the graph at train time.

\begin{table}[ht]
\caption{Characteristics of the datasets in our experiments}
\label{tab:datasets}
\centering
\begin{tabular}{lcccc}
\textbf{Dataset} & \textbf{Nodes} & \textbf{Edges} & \textbf{Classes} & \textbf{Features} \\
\midrule
Cora       & 2,708     & 5,429       & 7        & 1,433    \\
Citeseer   & 3,312     & 4,732       & 6        & 3,703    \\
Pubmed     & 19,717    & 44,338      & 3        & 500      \\
Reddit     & 232,965   & 11,606,919  & 41       & 602      \\
Amazon2M   & 2,449,029 & 61,859,140  & 47       & 100      \\
\bottomrule
\end{tabular}
\end{table}

\paragraph{Datasets}
The statistical properties of all datasets are summarized in Table \ref{tab:datasets}. The Cora \citep{cora}, Citeseer \citep{citeseer}, and Pubmed \citep{pubmed} datasets are citation network datasets where nodes represent documents and edges represent citation links. For the Cora and Citeseer datasets, node features are represented using a bag-of-words approach, while the Pubmed dataset uses Term Frequency-Inverse Document Frequency (TF-IDF) values. The Reddit dataset \citep{reddit} consists of posts from September 2014, with nodes representing individual posts and edges indicating interactions between posts by the same user. The Amazon2M dataset \citep{DBLP:journals/corr/abs-1905-07953} is a large-scale co-purchase network where nodes represent products and edges represent co-purchasing relationships.

\subsection{Results} \label{sec:results}

\paragraph{Cold-Start Classification.} 
The node classification performance on cold-start nodes, summarized in Table \ref{tab:classification}, demonstrates that our framework outperforms established baselines such as Cold-BREW, GraphSAGE and Graph-MLP\footnote{Graph-MLP did not provide code, and the results are as reported in \citep{zheng2021cold}.}. 

\begin{table*}[h]
\caption{Classification accuracy of cold-start nodes across several datasets. The top section lists existing baseline methods, while the bottom features our SPARC methods. Results highlighted in bold indicate the highest accuracy, while those underlined denote the second-highest accuracy in each category.}
\begin{center}
\begin{tabular}{lllllllll}
\bf METHOD 
&  \multicolumn{1}{c}{\bf Cora} & \multicolumn{1}{c}{\bf Citeseer} & \multicolumn{1}{c}{\bf Pubmed}  & \multicolumn{1}{c}{\bf Reddit}  & \multicolumn{1}{c}{\bf Amazon2M}  \\
\hline
G-SAGE
& 66.02 ± 1.18 &  51.46 ± 1.30 & 69.87 ± 1.10 & 85.63 ± 0.66 
 & OOM \\
C-BREW
& 68.92 ± 1.13 &  53.13 ± 0.24 & 72.32 ± 0.87  & OOM & OOM \\
Graph-MLP
& 65.00 &  53.82  & 71.22 & - & - \\
\hline
SPARC-GCN  & \textbf{73.88 ± 6.27} &  64.90 ± 3.19  & \underline{82.78 ± 2.05} &  \textbf{91.46 ± 0.92}  & \underline{72.33 ± 0.57} \\
SPARCphormer & 68.49 ± 0.89 & \underline{66.35 ± 0.92}  & \textbf{84.66 ± 0.25} & 90.70 ± 0.43 & \textbf{74.58 ± 0.19}\\
SPARC-Mamba & \underline{69.33 ± 3.96} & \textbf{70.10 ± 2.75} & 82.61 ± 0.65 & \underline{90.83 ± 0.26} & 72.02 ± 1.34 \\
\hline
\end{tabular}
\end{center}

\label{tab:classification}

\end{table*}

Unfortunately, we could not run G-SAGE and C-Brew on the Amazon2M dataset due to memory constraints\footnote{We were using the official implementation of these baselines.}.  Additionally, we were not able to run C-Brew also on the Reddit dataset. For this reason, there is no baseline for the Amazon2M dataset.
Nevertheless, we compare the performance of SPARC methods on cold-start nodes with that of baseline methods on test nodes. The results in Appendix \ref{sec:additional_results}, show that our SPARC methods remarkably demonstrate minimal performance degradation, with less than a 3\% drop in accuracy.

Moreover, the overall accuracy of SPARC architecture on test nodes remains comparable to state-of-the-art methods like Spectral-GCN and NAGphormer (see Appendix \ref{tab:classificationfull}).

\begin{figure*}[h]
    \centering  
    \includegraphics[scale=0.30]{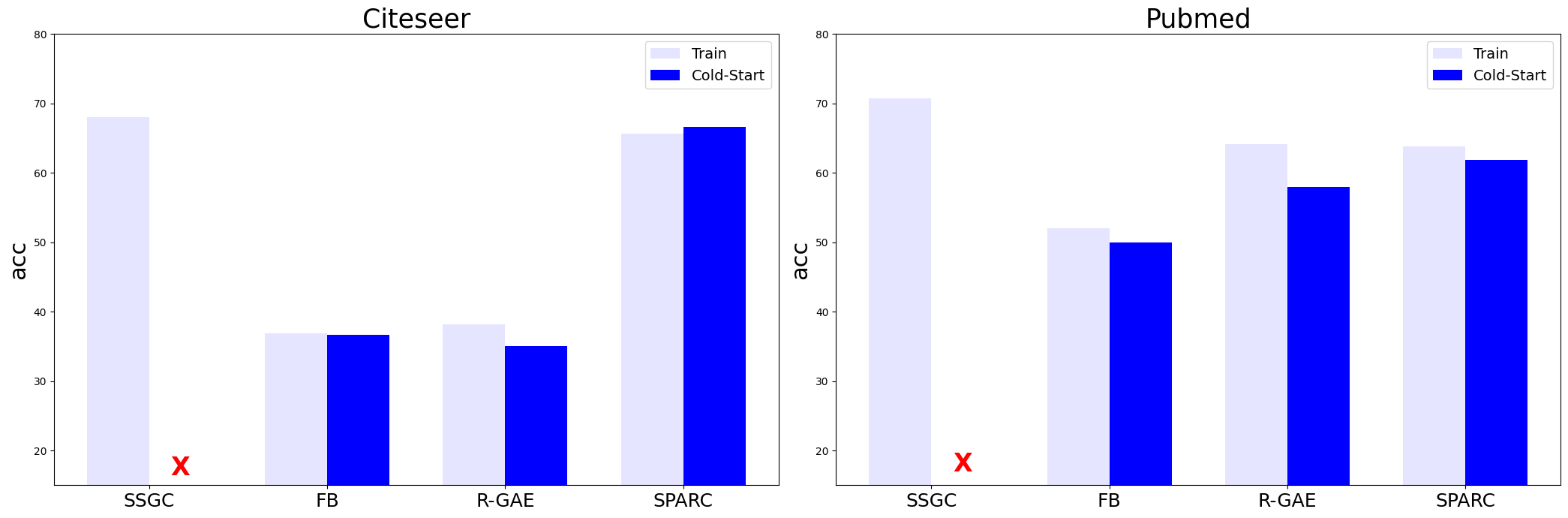}
    \label{fig}
    \caption{\textbf{Cold-start clustering accuracy.} Comparison of node clustering accuracies on the Citeseer and Pubmed datasets for both connected and cold-start nodes. The bar charts show accuracies for connected nodes (light blue bars) and cold-start nodes (blue bars). Marked with red crosses, SSGC fails to cluster cold-start nodes, while FB and R-GAE exhibit similar accuracy trends. Accuracy was measured using the Hungarian matching algorithm with node labels as ground truth.}
    \label{fig:clustering}
\end{figure*}

\paragraph{Cold-Start Clustering.} 
In our analysis of clustering techniques, we evaluated several methods, including feature-based (FB) method - $k$-means on the features, SSGC \citep{zhu2021simple}, and R-GAE \citep{mrabah2022rethinking}, across two benchmark datasets: Citeseer and Pubmed shown in Figure \ref{fig:clustering}.

SSGC performs best on connected nodes but does not apply to cold-start nodes. SPARC performs on-par with SSGC on connected nodes, while showing similar performance also on cold-start nodes. In addition, SPARC outperforms R-GAE and FB on both datasets, with a particularly large performance gap in Citeseer.

\paragraph{Cold-Start Link Prediction.} 
We measure cold-start link prediction accuracy by computing the mean reciprocal rank (MRR) \citep{guo2023linkless}. Specifically, we rank the nearest nodes in our embedding space to the cold-start node and compute the intersection with the actual nearest neighbors in the graph.

The results are presented in Table \ref{tab:link_prediction}. Link prediction for cold-start nodes is not well-researched; therefore, we compare SPARC to two approaches outlined in Linkless Link Prediction (LLP, \citet{guo2023linkless}). It utilizes relational knowledge distillation and cross-modeling of two networks: MLP and GNN.

As viewed in Table \ref{tab:link_prediction}, SPARC performs on-par with these methods. However, it is important to highlight that unlike more specialized models designed specifically for link prediction, SPARC does not require any additional architectural modifications or complex adjustments, making it a byproduct of our main approach. 

This inherent capability demonstrates the strength of our generalizable embedding in capturing the global structure of the graph, even when no direct connections exist for cold-start nodes. The comparable results, achieved with a simpler and more generalizable architecture, underscore the versatility of SPARC.

\begin{table}[h]
\caption{Link Prediction}
\label{tab:link_prediction}
\begin{center}
\scalebox{1}{
\begin{tabular}{lllll}
\bf METHOD & \multicolumn{1}{c}{\bf Cora} &  \multicolumn{1}{c}{\bf Citeseer} & \multicolumn{1}{c}{ \bf  Pubmed} \\ 
\hline
LLP-MLP  & 22.90 ± 2.22 & 28.21 ± 3.75 & 38.01 ± 1.67   \\
LLP&  27.87 ± 1.24 & 34.05 ± 2.45 & 50.48 ± 1.52 \\
\hline
SPARC & 31.25 ± 2.63 & 34.25 ± 2.78 & 48.07 ± 1.50  \\
\hline
\end{tabular}
}
\end{center}
\end{table}

\paragraph{SPARC for mini-batching}
\begin{figure}[h]
    \centering
    \includegraphics[width=0.6\linewidth]{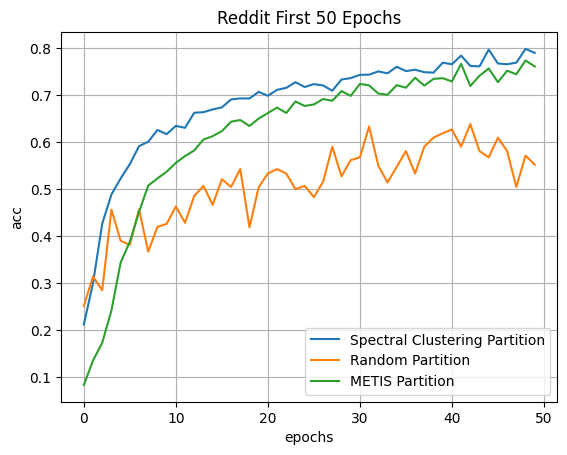}
    \caption{\textbf{Convergence rates on ClusterGCN and the Reddit dataset}. Training ClusterGCN on the Reddit dataset with three different mini-batching methods. The spectral clustered mini-batches result in faster convergence during the training process.}
    \label{fig:convergence}
\end{figure}

In this experiment, we train Cluster-GCN \citep{DBLP:journals/corr/abs-1905-07953} on Reddit with three mini-batching approaches, as follows. Randomly sampled mini-batches, METIS \citep{karypis1997metis} a graph partition algorithm as proposed in \citet{DBLP:journals/corr/abs-1905-07953} and spectral clustering using SPARC. We used each method to partition the graph into 1500 clusters and constructed mini-batches, each containing 20 clusters. Figure \ref{fig:convergence} shows that training with mini-batches constructed with SPARC demonstrates faster convergence in comparison to the other methods.

\paragraph{Analysis on varying amounts of cold-start nodes}
In this study, we assessed the efficacy of $\mathcal{F}_\theta$ embeddings in handling an increasing percentage of cold-start nodes, comparing SPARCphormer to two cold-start baselines- GraphSAGE and feature-based transformer on the Reddit dataset.

\begin{figure}[h]
    \centering
    \includegraphics[width=0.6\linewidth]{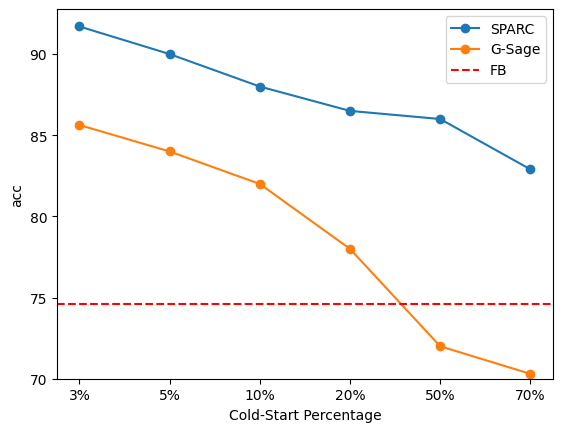}
    \caption{\textbf{Cold-start proportion analysis}: Accuracy trends of SPARCphormer, GraphSAGE, and the features-based (FB) transformer across increasing cold-start node proportion on the Reddit dataset. The FB transformer, which does not incorporate adjacency information and is oblivious to cold-start ratios, performs consistently at a lower accuracy level.}
    \label{fig:cold_start_precentage}
\end{figure}

To examine the impact of cold-start nodes, we systematically increased their proportion, effectively reducing the size of the training graph. As can be seen in Figure \ref{fig:cold_start_precentage}, SPARCphormer demonstrates slower performance degradation compared to Graph-SAGE as the proportion of cold-start nodes grows.



The FB transformer is similar to SPARCphormer except that the token lists are obtained via similarities in the feature space, which is inherently capable of performing inference on cold-start nodes. It does not suffer performance degradation, as it does not consider any adjacency information at all, which is also the reason for its lower performance, as explained in section \ref{motivation}. 
This highlights that the performance advantage of SPARCphormer stems primarily from the SPARC framework, rather than from the transformer architecture.
In addition, as can be seen, with more than 50\% cold-start nodes, the performance of GraphSAGE deteriorates below the level of the FB-transformer, while SPARC maintains a significantly higher performance. 

\section{Conclusion}

In this work, we introduced SPARC, a novel spectral-based method designed to address the cold-start problem in graph learning. SPARC allows state-of-the-art graph learning models to perform inference on cold-start nodes, a feature that many of them lack.
Key to this ability is the learning of a generalizable map into the eigenspace of the Laplacian operator on the underlying manifold.

We introduced SPARC as a general framework to incorporate this idea into any graph learning algorithm relying on adjacency information. In addition, we have presented three specific implementations of SPARC: SPARC-GCN, SPARCphormer, and SAMBA as well as applications for clustering, link prediction, and mini-batching.

Experimental results demonstrate that SPARC outperforms existing models in handling cold-start nodes, providing a solution for real-world applications where new nodes frequently appear. The adaptability of SPARC allows it to be seamlessly integrated into existing and future graph learning frameworks, enhancing its capability to manage evolving graphs. 

One caveat is that SPARC's reliance on the ability to learn the eigenfunction of the Laplacian from the input features, will not perform well when these features are of low quality. 

\paragraph{Impact Statement.} This paper presents work whose goal is to advance the field of Machine Learning. There are many potential social consequences of our work, none of which we feel should be specifically highlighted here.

\bibliography{references}
\bibliographystyle{plainnat}

\newpage
\appendix
\onecolumn

\section{Datasets}

\begin{table}[H]
\caption{Characteristics of the datasets in our experiments}
\label{tab:app:datasets}
\centering
\scalebox{1}{
    \begin{tabular}{lccccc}
    \textbf{Dataset} & \textbf{Nodes} & \textbf{Edges} & \textbf{Classes} & \textbf{Feats} & \textbf{Cold-Start \%} \\
    \hline
    Cora       & 2,708     & 5,429       & 7        & 1,433    & 3\% \\
    Citeseer   & 3,312     & 4,732       & 6        & 3,703    & 3\% \\
    Pubmed     & 19,717    & 44,338      & 3        & 500      & 3\% \\
    Reddit     & 232,965   & 11,606,919  & 41       & 602      & 3\% \\
    Amazon2M   & 2,449,029 & 61,859,140  & 47       & 100      & 3\% \\
    \hline
    \end{tabular}
}
\end{table}

\section{SPARC Prametric Map}
The generalization process borrows key ideas from SpectralNet \citep{shaham2018spectralnet} and spectral clustering to achieve a scalable and generalizable method for the first $k$ eigenvectors of the graph Laplacian. A key idea in spectral clustering is that embedding of the first $k$ eigenvectors (where $k \ll n$) captures the most significant variations in the graph structure. 

We computes the Laplacian matrix for a mini-batch using the graph adjacencies to find the parametric map using a neural network with orthogonal enforcement in the last layer. The training process is in a coordinated descent fashion, where we alternate between orthogonalization and gradient steps. Each of these steps uses a different mini-batch (possibly of different sizes), sampled from the training set $X$.

To guides the parametric map, we use the following Rayleigh-quotient loss: 
\begin{equation} \label{SNloss}
    \begin{aligned}
    \quad & \mathcal{L}_{RQ}= \text{trace} \left(Y^T{L}Y \right)  \quad  \text{s.t.}   \quad Y^TY = I
    \end{aligned}
\end{equation}
Where $Y$ is the network output and $L$ is the sub-Laplacian.  

\begin{figure}[H]
    \centering
    \includegraphics[width=\textwidth]{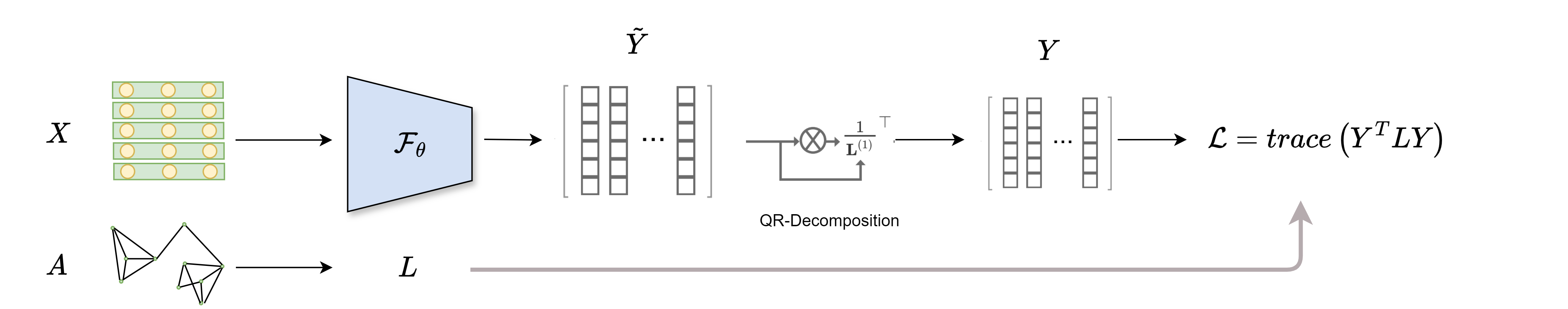}
    \caption{Learning a parametric map from the feature space $X$ to the first $k$ vectors of the Laplacian and undergoes QR decomposition
to ensure orthogonality}
    \label{fig:embedding-arch}
\end{figure}

The map learning process is detailed in Algorithm \ref{embedding-alg} and visually represented in Figure \ref{fig:embedding-arch}. The resulting embeddings correspond to the top $k$ eigenfunctions of the Laplacian matrix. Furthermore, the model is optimized to handle future nodes without incorporating any edge information during training.
\begin{algorithm}
\caption{$\mathcal{F}_\theta$ Training}
\begin{algorithmic}[1]
\REQUIRE $X \in \mathbb{R}^d$, number of vectors $k$, batch size $m$
\ENSURE Embedding $y_1, \ldots, y_n \in \mathbb{R}^k$
\STATE Part $X$ to mini-batches of size $m$ with neighbors 
\STATE Randomly initialize the network weights $\theta$
\WHILE{$\mathcal{L}_{RQ}$ not converged}
\STATE Orthogonalization step:
\STATE \quad Sample a mini-batch $X$ of size $m$ and the corresponding $A$ size $m\times m$;
\STATE \quad Forward propagate $X$ and compute inputs to orthogonalization layer $\hat{Y}$
\STATE \quad Compute the QR factorization $LL^T = \hat{Y}^T \hat{Y}$
\STATE \quad Set the weights of the orthogonalization layer to be $\sqrt{m} (L^{-1})^T$
\STATE Gradient step:
\STATE \quad Sample a mini-batch $X$ of size $m$ and the corresponding $A$;
\STATE \quad Forward propagate $x_1, \ldots, x_m$ to get $y_1, \ldots, y_m$
\STATE \quad Compute the loss $\mathcal{L}_{RQ}$
\STATE \quad Use the gradient of $\mathcal{L}_{RQ}$ to tune all $\mathcal{F}  $ weights, except those of the output layer
\ENDWHILE
\STATE Forward propagate $x_1, \ldots, x_n$ and obtain $\mathcal{F}_\theta$ outputs $y_1, \ldots, y_n$
\end{algorithmic}
\label{embedding-alg}
\end{algorithm}

\section{Exploring Different Laplacians}
Spectral eigendecompositions are typically performed solely on the graph structure, oblivious to any additional information associated with the graph. However, in most deep-learning models, different Laplacian matrices can yield better results for specific tasks to enhance unsupervised spectral embeddings. In our study, we explored two key approaches:

\paragraph{K-Power Random Walk.}
The k-power random walk method captures context from the k-hop neighbors within a graph by leveraging the k-power of the normalized adjacency matrix. The key idea lies in summing up these normalized matrices for each step, resulting in a new matrix denoted as $A_\text{k-power} = \sum_i ^k \text{normalized}(A^i)$, where $A$ is the graph adjacency matrix and normalization defined as $D^{-1/2}AD^{-1/2}$. This operation effectively simulates multiple convolutions, leading to smoother graphs with tighter clusters. In essence, the k-power random walk bridges local and global information, enhancing the expressive power of the graph.
\begin{figure}
    \centering
    \begin{minipage}{0.45\linewidth}
        \centering
        \includegraphics[width=\linewidth]{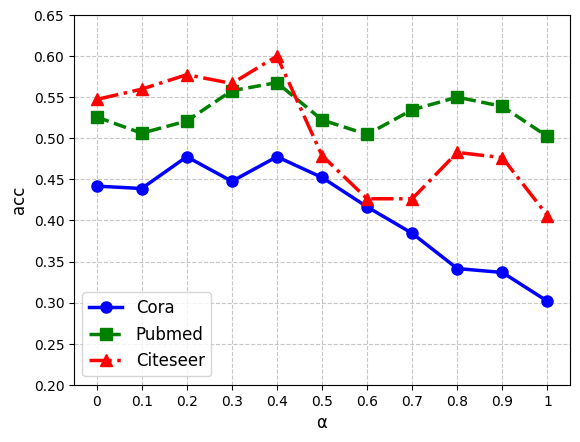}
        \caption{Clustering accuracy as a product of $\alpha$}
        \label{fig:ablation_alpha}
    \end{minipage}
    \hfill
    \begin{minipage}{0.45\linewidth}
        \centering
        \includegraphics[width=\linewidth]{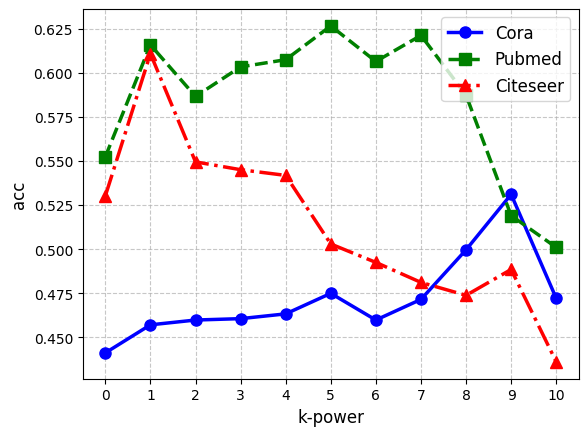}
        \caption{Clustering accuracy as a product of $k$}
        \label{fig:ablation_k}
    \end{minipage}
    \caption{Visualization of node clustering demonstrating the effects of over-smoothing and the effect of using solely feature affinities or adjacencies.}
    \label{fig:ablation-comparison}
\end{figure}

Figure \ref{fig:ablation_k} evaluates the effect of increasing the power $k$ of the adjacency matrix, which integrates increasingly distant neighborhood information into the graph representation. Initially, as $k$ increases, performance improves, reflecting the benefits of incorporating broader contextual information. However, beyond a certain point, further increases in $k$ lead to performance degradation, evident from the sharp declines for all datasets at higher $k$ values. This phenomenon, known as oversmoothing, occurs because the node representations begin to lose their distinctive characteristics, converging to a similar state that dilutes the useful signals for the learning tasks.

\paragraph{Features-Edges View.} \label{sec:feature_laplace}
In graph-based data, we construct simulated graphs with both node features and connections. In the context of node clustering, nodes may have defining features that indicate labels, neighborhood-defining labels, or a combination of both. Adapting ideas from traditional spectral clustering methods of constructing an affinity matrix based on the nodes' features. Given $n$ data points, an affinity matrix $W$ is an $n \times n$ matrix whose $W_{i,j}$ entry represents the similarity between $x_i$ and $x_j$. A popular choice for $W$ is the Gaussian kernel: $W_{i,j} = \exp \left(-\frac{\|x_i - x_j\|^2}{2\sigma ^2}\right)$ where $\sigma$ is a defined bandwidth. We construct the Laplacian matrix for a linear combination of the original adjacency matrix $A$ and the affinity matrix $W$: $A_\text{feat-edge} = \alpha * W + (1 - \alpha) * A$, where $\alpha$ depends on the characteristic of the nodes' features.

\section{Adotional Results}
\label{sec:additional_results}

\paragraph{Node Classification}
We present classification accuracy for two scenarios: 'Test', where nodes have full adjacency information, and 'Cold-Start', where connectivity data is missing. Our results show competitive performance with state-of-the-art models in the 'Test' scenario, while significantly outperforming existing methods in handling cold-start nodes. This highlights our method's unique capability to extend graph learning 
applications to effectively manage isolated nodes.

\begin{table}[H]
\caption{Classification}
\label{sample-table1}
\begin{center}
\scalebox{0.65}{
\begin{tabular}{lcccccccccc}
\bf METHOD & \multicolumn{2}{c}{\bf Cora} & \multicolumn{2}{c}{\bf Citeseer} & \multicolumn{2}{c}{\bf Pubmed} & \multicolumn{2}{c}{\bf Reddit} & \multicolumn{2}{c}{\bf Amazon2M} \\
\cline{2-11} 
& \multicolumn{1}{c}{\bf Test} & \multicolumn{1}{c}{\bf Cold-Start} & \multicolumn{1}{c}{\bf Test} & \multicolumn{1}{c}{\bf Cold-Start} & \multicolumn{1}{c}{\bf Test} & \multicolumn{1}{c}{\bf Cold-Start} & \multicolumn{1}{c}{\bf Test} & \multicolumn{1}{c}{\bf Cold-Start} & \multicolumn{1}{c}{\bf Test} & \multicolumn{1}{c}{\bf Cold-Start} \\
\hline
Spectral-GCN & 87.94 ± 0.85 & NS & 77.92 ± 0.61 & NS & 86.20 ± 0.41 & NS & OOM & NS & OOM & NS \\
NAGphoremer & 89.55 ± 0.48 & NS & 76.32 ± 0.52 & NS & 88.30 ± 0.29 & NS & 93.75 ± 0.03 & NS & 77.43 ± 0.24 & NS \\
\hline
G-SAGE & 83.92 ± 1.25 & 66.02 ± 1.18 & 71.78 ± 2.67 & 51.46 ± 1.30 & 82.16 ± 1.92 & 69.87 ± 1.10 & 94.32 ± 0.00 & 85.63 ± 0.66 & OOM & OOM \\
C-BREW & 84.66 ± 0.00 & 69.62 ± 0.00 & 71.18 ± 0.00 & 53.17 ± 0.00 & 86.81 ± 0.00 & 72.33 ± 0.00 & OOM & OOM & OOM & OOM \\
Graph-MLP & - & 65.00 & - & 53.82 & - &  71.22 & - & - & - & - \\
\hline
SPARC-GCN & 84.46 ± 1.51 & 73.88 ± 6.27 & 69.44 ± 4.23 & 63.66 ± 0.00 & 86.46 ± 3.77 & 82.78 ± 2.05 & 93.04 ± 0.87 & 91.46 ± 0.92 & 73.83 ± 0.20 & 72.33 ± 0.57 \\
SPARCphormer & 79.41 ± 1.65 & 68.49 ± 0.89 & 73.80 ± 0.43 & 66.35 ± 0.92 & 85.12 ± 0.46 & 84.66 ± 0.25 & 94.67 ± 0.76 & 90.70 ± 0.43 & 74.58 ± 0.19 & 74.58 ± 0.19 \\
SAMBA & 81.48 ± 1.12 & 69.33 ± 3.96 & 71.99 ± 0.71 & 70.10 ± 2.75 & 86.40 ± 0.90 & 82.61 ± 0.65 & 93.39 ± 0.23 & 90.83 ± 0.26 & 72.72 ± 0.33  & 72.02 ± 1.34 \\
\hline

\end{tabular}}
\end{center}
\label{tab:classificationfull}
\end{table}

\paragraph{Neighborhood Prediction} \label{sec:neighborhood_prediction}
To continue the analysis presented in Figure \ref{fig:neighborhood_prediction_accuracy}, which compared the efficacy of spectral embedding space versus feature space in predicting neighborhood, we extend our examination to include the $\mathcal{F}_\theta$ embedding space. 

Although the prediction accuracy in the $\mathcal{F}_\theta$
space are lower then that achieved with spectral embeddings, this is largely due to the spectral embeddings being directly constructed from adjacency information through the Laplacian matrix, whereas $\mathcal{F}_\theta$ learns the mapping without exposure to cold-start nodes. Nevertheless, $\mathcal{F}_\theta$
substantially outperforms the naive feature space approach, highlighting our method's significant advances in effectively handling new cold-start nodes and affirming its practical utility and robustness.
  
\begin{table}[h]
\centering
\caption{Comparison of various spaces in neighborhood prediction}
\label{tab:neighborhood_prediction}
\scalebox{1}{
\begin{tabular}{@{}lcccccc@{}}
\toprule

Method & Cora & Citeseer & PubMed & Reddit \\ 
\midrule
Feature Based       & 35.78\% & 3.90\% & 26.68\% & 34.35\% \\
Spectral Embedding  &  83.86\% & 70.21\% & 79.32\% &  70.02\% \\
$\mathcal{F}_\theta$ Embedding     & 62.81\% & 58.76\% & 58.58\% & 68.66\% \\
\bottomrule
\end{tabular}
}
\end{table}

\section{Algorithms}

\begin{algorithm}[H]
\caption{SPARC-GCN Training}
\begin{algorithmic}[1]
\REQUIRE Node features $X \in \mathbb{R}^{n \times d}$, batch size $m$, labels $\mathbf{y}$ for training nodes
\ENSURE Predicted labels $\hat{\mathbf{y}}_1, \ldots, \hat{\mathbf{y}}_n$
\STATE Compute spectral embeddings $U$ for all nodes in $X$ (generalizable to cold-start nodes)
\STATE Randomly initialize model parameters $\theta$
\WHILE{$L_{\text{Spectral-GCN}}(\theta)$ not converged}
    \STATE Sample a mini-batch of near-neighbors in the spectral space $U$
    \STATE Forward propagate Spectral-GCN-Layers and Linear Layers get final embeddings $z_1, \ldots, z_m$
    \STATE Apply a linear layer to predict labels $\hat{\mathbf{y}}_1, \ldots, \hat{\mathbf{y}}_m$
    \STATE Compute the loss $L_{\text{Spectral-GCN}}(\theta)$ using the ground-truth labels $\mathbf{y}$
    \STATE Update the model parameters $\theta$ using gradient descent
\ENDWHILE
\STATE \textbf{Cold-Start Inference:} $\hat{x}$
    \STATE Predict $\hat{U}$ spectral embedding of the cold-start node
    \STATE Identify the $k$ nearest neighbors of $\hat{x}$ in the spectral embedding space $U$
    \STATE Forward propagating the new inference batch
\end{algorithmic}
\label{alg:spectral-gcn}
\end{algorithm}

\begin{algorithm}[H]
\caption{SPARCphormer Training}
\begin{algorithmic}[1]
\REQUIRE Node features $X \in \mathbb{R}^{n \times d}$, number of neighbors $k$, labels $\mathbf{y}$ for training nodes
\ENSURE Predicted labels $\hat{\mathbf{y}}_1, \ldots, \hat{\mathbf{y}}_n$
\STATE Compute our SPARC embeddings for all nodes in $X$ 
\FOR{each node $v \in \mathcal{V}$}
    \STATE Identify the $2^k$ nearest neighbors of $v$ in the spectral embedding space
    \STATE Construct token list for $v$ using features of the $2^k$ nearest neighbors as described in \ref{sec:SPARCphormer}
\ENDFOR
\STATE Randomly initialize model parameters $\theta$
\WHILE{$\mathcal{L}_{\text{nll}}(\theta)$ not converged}
    \STATE Sample a mini-batch of training nodes and their token lists
    \STATE Forward propagate token lists through the self-attention mechanism to get final embeddings $z_1, \ldots, z_m$
    \STATE Apply a linear layer to predict labels $\hat{\mathbf{y}}_1, \ldots, \hat{\mathbf{y}}_m$
    \STATE Compute the loss $\mathcal{L}_{\text{nll}}(\theta)$ using the ground-truth labels $\mathbf{y}$
    \STATE Update the model parameters $\theta$ using gradient descent
\ENDWHILE
\end{algorithmic}
\label{alg:spectral-graphormer}
\end{algorithm}

\begin{algorithm}[H]
\caption{SAMBA Training}
\begin{algorithmic}[1]
\REQUIRE Node features $X \in \mathbb{R}^{n \times d}$, number of neighbors $k$, labels $\mathbf{y}$ for training nodes
\ENSURE Predicted labels $\hat{\mathbf{y}}_1, \ldots, \hat{\mathbf{y}}_n$
\STATE Compute our SPARC embeddings for all nodes in $X$ 
\FOR{each node $v \in \mathcal{V}$}
    \STATE Identify the $2^k$ nearest neighbors of $v$ in the spectral embedding space
    \STATE Construct token list for $v$ using features of the $2^k$ nearest neighbors as described in \ref{sec:SPARCphormer}
\ENDFOR
\STATE Randomly initialize model parameters $\theta$
\WHILE{$\mathcal{L}_{\text{nll}}(\theta)$ not converged}
    \STATE Sample a mini-batch of training nodes and their token lists
    \STATE Forward propagate token lists through the MAMBA mechanism to get final embeddings $z_1, \ldots, z_m$
    \STATE Apply a linear layer to predict labels $\hat{\mathbf{y}}_1, \ldots, \hat{\mathbf{y}}_m$
    \STATE Compute the loss $\mathcal{L}_{\text{nll}}(\theta)$ using the ground-truth labels $\mathbf{y}$
    \STATE Update the model parameters $\theta$ using gradient descent
\ENDWHILE
\end{algorithmic}
\label{alg:spectral-graphormer}
\end{algorithm}

\section{Technical Detail and Hyper-parameters.}
For fairness, we run each of the compared algorithms ten times on the above datasets, recording both the mean and standard deviation of their performance. The same backbones are employed across all methods and datasets.
All external algorithms provided hyper-parameters and so each run consists of the reported parameters.

\begin{table}[H]
\centering
\caption{Hyperparameters for SPARC-Embeddings}
\label{tab:hyperparameters}
\begin{tabular}{lcccc}
\hline
\textbf{Parameter} & \textbf{Cora} & \textbf{Citeseer} & \textbf{Pubmed} & \textbf{Reddit} \\ \hline
Hidden Dimension   & 512, 256, 32           & 512, 256, 32              & 512, 256, 32             & 512, 256, 64             \\
K eigenvectors    & 32            & 32               & 32              & 64              \\
Peak Learning Rate & 0.1         & 0.1            & 0.1           & 0.1           \\
Weight Decay       & $1e-5$        & $1e-5$           & $1e-5$          & $1e-5$          \\
\hline
\end{tabular}
\end{table}

\begin{table}[H]
\centering
\caption{Hyperparameters for SPARC-GCN}
\label{tab:hyperparameters}
\begin{tabular}{lcccc}
\hline
\textbf{Parameter} & \textbf{Cora} & \textbf{Citeseer} & \textbf{Pubmed} & \textbf{Reddit} \\ \hline
Dropout            & 0.1           & 0.1              & 0.1             & 0.1             \\
Hidden Dimension   & 64, 256, 7           & 64, 256, 6              & 64,256, 3             & 64, 256, 41             \\
Peak Learning Rate & 0.1         & 0.1            & 0.1           & 0.1           \\
Weight Decay       & $1e-5$        & $1e-5$           & $1e-5$          & $1e-5$          \\
\hline
\end{tabular}
\end{table}

\begin{table}[H]
\centering
\caption{Hyperparameters for SPARCphormer and SAMBA}
\label{tab:hyperparameters}
\begin{tabular}{lcccc}
\hline
\textbf{Parameter} & \textbf{Cora} & \textbf{Citeseer} & \textbf{Pubmed} & \textbf{Reddit} \\ \hline
Dropout            & 0.1           & 0.1              & 0.1             & 0.1             \\
Hidden Dimension   & 512           & 512              & 512             & 512             \\
Token List Size    & 5            & 7               & 10              & 13              \\
Number of Heads    & 8             & 8                & 8               & 8               \\
Peak Learning Rate & 0.001         & 0.001            & 0.001           & 0.001           \\
Weight Decay       & $1e-5$        & $1e-5$           & $1e-5$          & $1e-5$          \\
\hline
\end{tabular}
\end{table}

\section{OS and Hardware}
The training procedures were executed on Rocky Linux 9.3, utilizing Nvidia 578 GPUs including GeForce GTX 1080 Ti and A100 80GB PCIe.

\end{document}